\documentclass{article}
\usepackage{ijcai23}

\usepackage{times}
\usepackage{soul}
\usepackage{url}
\usepackage[hidelinks]{hyperref}
\usepackage[utf8]{inputenc}
\usepackage[small]{caption}
\usepackage{graphicx}
\usepackage{amsmath}
\usepackage{amsthm}
\usepackage{booktabs}
\usepackage{algorithm}
\usepackage{algorithmic}
\usepackage[switch]{lineno}

\usepackage{xcolor}
\usepackage{amsfonts}
\usepackage{array, comment}
\usepackage{mdframed}
\usepackage{lipsum}

\usepackage{enumitem}

\title{Topological Deep Learning: A Review of an Emerging Paradigm}

\author{
Ali Zia$^{1,2}$
\and
Abdelwahed Khamis$^2$\and
James Nichols$^1$\and
Zeeshan Hayder$^2$\and
Vivien Rolland$^2$\And
Lars Petersson$^2$
\affiliations
$^1$College of Science, Australian National University, Australia\\
$^2$CSIRO, Australia
 }

\begin{document}

\maketitle

\begin{abstract}
    {\em Topological data analysis} (TDA) provides insight into data shape. The summaries obtained by these methods are principled global descriptions of multi-dimensional data whilst exhibiting stable properties such as robustness to deformation and noise. Such properties are desirable in deep learning pipelines but they are typically obtained using non-TDA strategies. This is partly caused by the difficulty of combining TDA constructs (e.g. barcode and persistence diagrams) with current deep learning algorithms. Fortunately, we are now witnessing a growth of deep learning applications embracing topologically-guided components. 
    In this survey, we review the nascent field of {\em topological deep learning} by first revisiting core concepts of TDA. We then explore how the use of TDA techniques has evolved over time to support deep learning frameworks, and how they can be integrated into different aspects of deep learning. Furthermore, we touch on TDA usage for analyzing existing deep models; {\em deep topological analytics}.
    Finally, we discuss the challenges and future prospects of topological deep learning.
\end{abstract}



\begin{figure*}[t!]
    \centering
    \includegraphics[width= 1\linewidth]{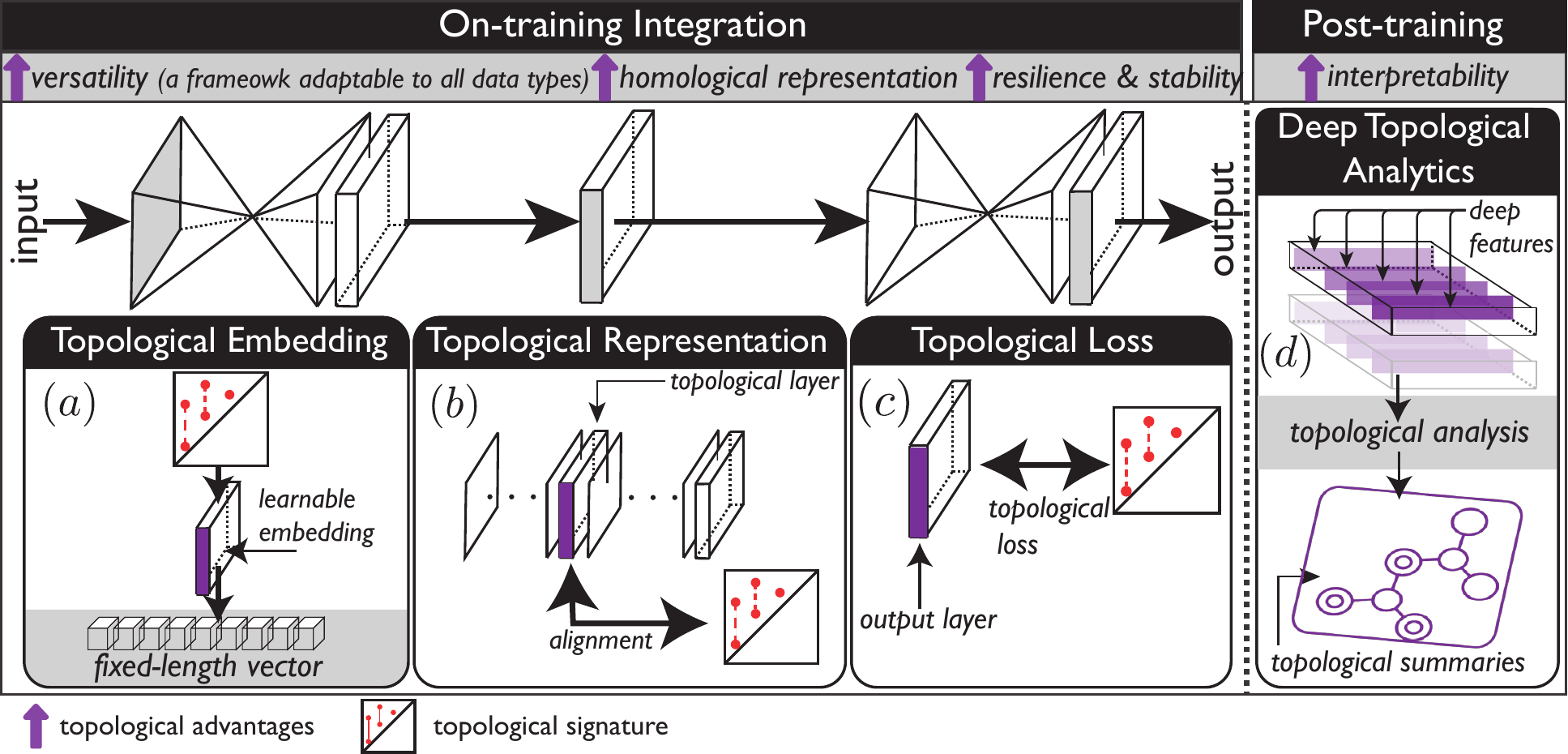}
    \caption{\textbf{Topological Deep Learning} introduces TDA methods to deep models leading to topological neural architectures that can potentially address deep learning limitations. This is done by plugging topological components for (a) learning  features \textbf{Embedding} (Section~\ref{sec:tda_embedding}), (b) enhancing the learned \textbf{Representations} (Section~\ref{sec:tda_integration}), and/or (c)  regularizing the model using a topological \textbf{Loss} (Section~\ref{sec:top_loss}). Beyond that, (d) TDA can be used \textit{post-training} to reveal insights of trained models (interpretability) (Section~\ref{sec:tda-measure}).}
    \label{fig:taxonomy}
\end{figure*}     
    

\section{Introduction}

Topological data analysis (TDA) is a relatively recent amalgam of theory and algorithms that aim to obtain a geometric and topological understanding of data from real world applications. 
The approach to data employed in TDA fundamentally differs from that in statistical learning. Rather than finding summary statistics, estimators, fitting approximate distributions, clustering, or training neural nets, TDA instead seeks to understand the properties of the geometric object, often a \emph{manifold}, on which the data resides. This reflects the common intuition that data tends to lie on, or close to, a lower dimensional manifold that is embedded in high dimensional feature space. In this article, we sometimes refer to this as the \emph{data manifold}.

The main goal of TDA is to infer information about the global structure of the data manifold, such as its connectivity and the presence of multi-dimensional holes.
An important property of the topological information obtained is its invariance to continuous deformation and scaling. This property also lends itself to robustness against perturbation and noise. 
Another benefit is the versatility of the TDA methods, owed mostly to the abstract origins of algebraic topology. The methods are applicable to a wide variety of data types and objects. This includes point cloud data in Euclidean spaces, categorical data, or the analysis of images and functions.
Due to these aspects, the absence of parameters to tune, and the fundamental mathematical nature of the TDA approach, it is intriguing to include it in deep neural networks.

There has been much recent activity in co-opting topological approaches in deep learning, however, there remain considerable open questions as to what the leading approach should be, due to many computational and theoretical concerns.
The TDA methods discussed in this paper form but a small part of the ever-expanding interface between topological data analysis and machine learning. We did our best to choose work that has a historical and linear connection with deep learning approaches, to improve understandability.


This survey provides the broader machine learning community with a convenient starting point to explore how TDA has been integrated with deep learning. To the best of our knowledge, this is the first work that comprehensively covers topological deep learning and organizes the research works in this field in a unified taxonomy (Section~\ref{sec:deep_tda}).

We start in Section \ref{sec:overview} by introducing the key theoretical concepts of TDA and their representations for learning. In Section \ref{sec:deep_tda} we explain how topological approaches can fit into different deep learning constructs, such as learnable features, feature transformations, and loss functions. In Section \ref{sec:tda-measure} we shed the light on a promising use of TDA to understand and dissect trained deep models, called \textit{deep topological analytics}.


 
We continue in Section \ref{sec:challenges} with a discussion of the known challenges of TDA and its adaptation to deep neural networks. We further discuss future directions and adjacent applications of topological deep learning, and we present some current libraries. Finally, we make some concluding remarks in Section \ref{sec:conclusion}.

\textbf{Notations:} We write $X \in \mathbb{R}^{n \times d}$ to denote the data set, where $n$ is the number of samples and $d$ the number of features or dimensions. 
We write $\mathcal{M}$ to denote the underlying data manifold, which for the purposes of this survey is a locally Euclidean space embedded in $\mathbb{R}^d$.
We write BD and PD as abbreviations of {\em barcode diagram} and {\em persistence diagram}.


\section{Overview of TDA} \label{sec:overview}


An object's {\em topology} is broadly defined as the characteristics that remain invariant under continuous deformation, as if the object was made of soft rubber. 
How many connected components the object contains, the holes or voids it contains, and how the object loops back on itself are a few examples of topological properties.
In a sense, topological information can be considered qualitative. For example, if we demonstrate that data points lie on two totally disconnected sub-manifolds, then we know that the data comes from two very distinct sources, or that the underlying system has two distinct states.

A central concept is that of {\em homology}, which is a powerful tool to characterize the topological features of a space.
Homology is an abstract concept. It is difficult to work with, and its general definition is outside the scope of this paper or even most of the TDA literature \cite{carlssonTopologyData2009}. 
In essence, the $k$-th homology (where $0\le k \le d$) is a group that characterizes the set of $k$-dimensional holes (or voids) in a topological space.
A 1-dimensional hole can be traced around with a 1-dimensional loop (like a loop of string), whereas a 2-dimensional hole is a void, for example, the void within a hollow sphere in 3 dimensions.
These $k$-dimensional holes are counted by the {\em Betti numbers}. The $k$-th Betti number is defined as the rank of the $k$-th homology, which in general can be quite difficult to compute.
Fortunately, there are some spaces for which the Betti numbers are relatively straightforward to compute.





\subsection{Simplicial complexes and persistent homology}

The $k$-th homology is much more convenient to work with when we restrict ourselves to {\em simplicial complexes}, which are structures built upon discrete sets.
This is the natural domain for data-driven and machine learning applications.



A {\em simplex} can be considered a generalization of a triangle or tetrahedron, it is the simplest polytope of any given dimension.
A simplex in zero dimensions is a point, in one dimension is a line segment, in two dimensions is a triangle, in three is a tetrahedron, and so on. 
We use {\em $k$-simplex} to refer to a simplex of dimension $k$.
Note that any simplex is composed of {\em faces} which are themselves simplices of lower dimension.
A {\em simplicial complex} $K$ is a collection of simplices with two properties: each face of a simplex in $K$ must also be in $K$, and the intersection of any two simplices of $K$ is either empty or a face of both of them.

Consider each point in our data set $X$ to be a vertex (a 0-simplex). We can define a set of 1-simplices as connections between pairs of vertices, 2-simplices between collections of three vertices, and so on. Thus we build a simplicial complex $K$ that gives some sense of ``connectivity'' between data points.
It can be thought of as a {\em hyper-graph} on $X$.
Note that $K$ is not necessarily unique on $X$. 

Homological information is much easier obtained for a simplicial complex, and in particular, the $k$-th Betti number can be obtained through tractable linear algebra.
The Betti numbers in this setting are closely related to {\em Euler characteristic}, which gives the relationship between the numbers of vertices, edges, and faces in a polyhedron. 

The goal now is to construct simplicial complexes on $X$ that reflect the underlying topology of $\mathcal{M}$. This is done by varying scale, typically a radius $r>0$.
The \v{C}ech complex and the Vietoris-Rips complex are two typical constructions \cite{chazal2021introduction}. A \v{C}ech complex $C_r(X)$ includes a $k$-simplex on $(k + 1)$ vertices of $X$ if the collection of balls of radius $r$ centered on each vertex has a non-empty intersection.
The Vietoris-Rips (or simply Rips) complex $V_r(X)$ includes a $k$-simplex on any set of $(k + 1)$ vertices that  all have a pairwise distance less than $r$ of each other \cite{Zomorodian2010}. These two constructions of simplicial complexes can yield very different results on the same data set with the same $r$. 

{\em Persistent homology} is obtained through a {\em filtration} $F$, which is a growing sequence of sub-complexes: $K_1 \subseteq K_2  \subseteq \ldots \subseteq K_n = K$.
Two commonly used examples of filtration are the sets of simplicial complexes, $C_r(X)$ or $V_r(X)$, that are obtained with increasing radius $r$. 
As we vary $r$, these constructs will naturally reflect different aspects of the topology of $\mathcal{M}$.
There is monotone inclusion of these simplicial complexes with increasing $r$, i.e.~for two radii $r \le r^\prime$ we have that $C_r(X) \subseteq C_{r^\prime}(X)$ and $V_r(X) \subseteq V_{r^\prime}(X)$. 


The key idea is to track changes in topological features as they appear and disappear over the filtration. 
We may see new loops created, separate components connected, or holes filled in as we increase $r$. 
We record the lifetime of these features with respect to $r$, that is the \textit{appearance} (at $b_i$ for birth) and \textit{disappearance} (at $d_i$ for death) of a particular topological feature. 






\subsection{Representations of persistent homology}
\label{sec:homology_representation}

The set of birth and death coordinates obtained from the filtration forms the backbone of persistent homology. 
The two most popular representations of this information are {\em barcode diagrams} and {\em persistence diagrams} \cite{carlssonTopologyData2009}.
The multi-set of intervals $(b_i, d_i)$ form the barcode diagram (BD), the name coming from the visual representation of the set of intervals as stacked line segments.
In the persistence diagram (PD) the lifetime of each feature is represented by a point in $\mathbb{R}^2$ with coordinates $(b_i, d_i)$. 
A filtration may have several copies of the same birth and death interval, which is represented in the PD by giving the point $(b_i, d_i)$ an integer valued multiplicity. 
It is important to note that the BD and PD contain equivalent information and 
one can define a bijection between the two. From here onwards we use the term PD to refer to either construct unless BD is explicitly referred to.






A data set's PD contains a wealth of topological information.
Features that have a long persistence interval ($d_i - b_i$) are considered to be likely to reflect the true topological features of the underlying manifold $\mathcal{M}$. 
These features are represented in the PD by points far away from the diagonal.
A short persistence interval describes a feature that is possibly generated from noise or is otherwise insignificant.
Features with short persistence  will be represented by points close to the diagonal line in the PD.
Hence, points in the PD further from the diagonal are considered more informative.

Comparing the PDs of two objects is a way to assess their topological similarity. In the next section, we discuss various methods to represent them in manners suitable for machine learning and computation.

\subsection{Homological feature vectorizations}
\label{sec:representation}
         



Most machine learning methods assume that the input data resides in $\mathbb{R}^d$ or more generally some Hilbert space $\mathcal{H}$. Hence they cannot be directly applied to datasets comprised of PDs, and the multi-set information contained in the PD needs to be represented in some vector format. This process is called {\em vectorization}, which requires the definition of a continuous map $f: \mathrm{PD} \to \mathcal{H}$.
There is a plethora of different methods in the literature and there are some subtle consequences that come with different choices of vectorization techniques \cite{aliSurveyVectorizationMethods2022}.
It is important to note that these vectorization methods can be thought of as handcrafted feature engineering, rather than feature learning. 
In this section, we discuss various strategies that have evolved over time. 

         


       A simple approach for representing PDs is using their {\em statistical properties} such as the sum, mean, variance, maximum, minimum, etc \cite{aliSurveyVectorizationMethods2022}.
       The total Betti number of a certain filtration can also be used as a summary representation \cite{Cang2015}. These approaches yield a univariate output and lose information, however can still be useful. 
       

        Another approach is to vectorize BDs using {\em histogram-like methods} \cite{Cang2017}.
        The basic concept is to discretize the BD along the filtration axis, creating equal sized bins in which we count the number of persistent intervals. 
       Alternatively, {\em tropical coordinates} defined on the space of BDs are a useful and stable algebraic representation \cite{Kalisnik2018}. 

        Yet a different approach is to construct various forms of {\em persistence functions} from PDs. 
        These functions are readily vectorized themselves, however, it is also convenient to work with them directly for many tasks \cite{Bubenik2020,Adams2017}.
        Example of these persistent functions includes persistence landscape \cite{Bubenik2020}
        , persistence Betti number \cite{edelsbrunnerTopologicalPersistenceSimplification2002}, persistence Betti function \cite{Xia2017}, persistence surfaces and persistence images \cite{Adams2017}, etc. 
        
        
        A useful feature representation technique called {\em persistence codebooks} \cite{Zielinski2020} uses bag-of-words quantization techniques 
        to group data points into a fixed sized vector. Chevyrev et. al. \cite{Chevyrev2020} proposed persistence paths, which is a feature map for barcodes. 

        Representation can vary from simple to complex structures. 
        To get better structural representations there is scope to investigate new methods of vectorization which can benefit topological learning models.
        Note however that when a large feature vector is used to represent PDs, the curse of dimensionality comes into play. 
        In this case, variable selection, regularization approaches, or dropout methods should be considered \cite{Pun2022}.
        
        In addition, it is important to consider the comparison of different PDs. To this end various metrics have been proposed, such as
        bottleneck distance \cite{Mileyko2011}, and adaptations of Gromov-Hausdorff and Wasserstein metric \cite{Bubenik2017}. Many other metrics have been considered in the literature as well. A central consideration is the {\em stability} of vectorizations and metrics. 
        We discuss this further in Section \ref{sec:top_loss}.
         



        As discussed, vectorization methods can be used in input space, however, 
        kernel-based models are another important way to combine PD information with machine learning models \cite{Kwitt2015}.  
        Since metrics can be modified into kernels, various approaches have been proposed to induce kernel function from PD information~\cite{Pun2022}
        and into traditional machine learning approaches like PCA and SVM. 
        Topological-based kernel methods have been used successfully in various ways \cite{Zhu2016,Kwitt2015}, however techniques based on kernel methods suffer from scalability issues~\cite{Pun2022}, as training typically scales poorly with the sample number (e.g., roughly cubic in the case of kernel-SVMs). 
        We do not discuss topological kernel methods any further in this paper.

         Many of the aforementioned methods have advantageous stability properties with respect to standard metrics in TDA like the Wasserstein or Bottleneck distances. However, they all have the same drawback: the mapping of topological representation that is compatible with existing learning techniques is pre-defined. Therefore, it is fixed and agnostic to any specific learning task, which makes it suboptimal. The phenomenal success of deep neural networks has shown that learning representations (i.e.~feature learning) is a preferable approach.



\section{Topological Deep Learning (TDL)} 
\label{sec:deep_tda}



Topological representations that incorporate structural information hold great promise for topological deep learning models \cite{hofer2017deep}.
Combining these cues with deep learning approaches has inherent benefits in various applications. On the flip side, deep learning approaches can be useful in overcoming some common hurdles faced by TDA approaches in estimating robust topological features. 
The incorporation of topological concepts into deep learning has only recently been investigated and the following are general benefits:






\setlist[itemize]{leftmargin=*}

\begin{itemize}

\item {\em Global features} from input data can be efficiently and robustly extracted that would otherwise be inaccessible via traditional feature maps.

\item TDA is {\em versatile and adaptable}, meaning that we are not limited to specific problems and types of data (such as images, sensor measurements, time series, graphs, etc.).

\item TDA is {\em noise-resistant} in several different problems, including the classification of 3D surface meshes, the recognition of 2D object shapes, the manifold of natural image patches, analyzing activity patterns of the visual cortex, and clustering~\cite{Pun2022,aliSurveyVectorizationMethods2022}. 

\item TDA can be applied to {\em arbitrary data structures} even without any prepossessing, with the right filtrations.

\item A new trend is emerging that allows {\em efficient back-propagation} through persistent homology components. This is a long-standing challenge in TDA (further discussed in Sec.~\ref{sec:top_loss}), but now topological layers are becoming compatible with deep learning and end-to-end training schemes.


\end{itemize}

We reiterate that though the benefits of using TDA (more specifically persistent homology) and deep learning together have demonstrated success, there are still some theoretical and computational challenges in the application of TDA to data. We discuss these issues at length in Section \ref{sec:challenges}.




In the rest of this section, we investigate TDA for deep learning from lenses of different magnifications and perspectives as shown in Figure~\ref{fig:taxonomy}. In particular, we explore the use of persistent homology in various different ways. The discussion in Section~\ref{sec:tda_embedding}-\ref{sec:top_loss} is focused on the \textit{on-training integration} of TDA. That is, building topological neural architectures. However, a holistic view should also consider TDA's contribution to \textit{post-training} (deep topological analytics). This uses TDA to study the `shape' of a trained model. Thus, we review works that studied deep model complexity and interpretability using TDA in Section~\ref{sec:tda-measure}. 



\subsection{Learning Topological Features Embedding}\label{sec:tda_embedding}

    In this section, we extend the discussion of fixed vectorization methods (Section~\ref{sec:representation} ) by introducing deep learnable vectorization (i.e. embedding). A key advantage here is the possibility of leveraging the deep model to simultaneously learn the vectorization of data and the representation of the target task. For example,  we may parameterize the vectorization of persistence diagrams $\mathrm{PD}$ to embedding vector $V \in \mathbb{R}^d$ by neural layers $f_w$ where $w$ denotes the trainable parameters. Guided by the task loss,  we can efficiently learn mapping $f_w: \mathrm{PD_x} \to V_{x}$ and automatically answer the question of ``which family of vectorizations should best work for the given task''.

    Handling PDs by neural networks is the focus of many deep topological embedding works. Generally, PDs deep vectorization layers should be continuous and permutation invariant with respect to the input. The latter requirement is motivated by the {\em set} nature of the persistence diagram.  Hofer \emph{et al.}~\cite{hofer2017deep} introduced the first learnable deep vectorization of PDs. It adopts a permutation invariant transformation by evaluating the PD's points against Guassian(s) whose mean and variance are learned during the training. Since permutation invariance was explored in other deep learning problems (e.g. Deep Set \cite{Zaheer2017} for points cloud), some vectorization techniques for PD were borrowed from them. For example, PersLay~\cite{Carriere2019} builds on DeepSets for embedding extended PDs encoding graphs and uses it for graph classification. Recently, transformers were used for PDs embedding. Persformer~\cite{Reinauer2021} architecture showed superiority in synthetic and graph tasks while having some interpretability features. Note that transformers without positional encoding can be made as expressive as Deep Sets. Thus, the permutation invariance requirement can be maintained.

    Beyond PDs, deep embedding was explored for other topological signatures. For example, PLLay~\cite{Kim2020} provides a layer for embedding persistence landscapes. PLLay claim to robustness to extreme topological distortion is backed by a tight stability bound that's independent of the input complexity.




     

Topological embedding transforms the topological input with a complex structure into a vector representation compatible with deep models. As discussed in this section, the process uses a custom topological input layer for embedding. In the next section, we explore topological components that enhance deep learning representation and usually have the flexibility to be plugged anywhere in the network.

\subsection{Integration of Topological Representations}\label{sec:tda_integration}

    Representation learning is the process of learning features from data that can be used to improve the accuracy of the model. Deep learning excels in this regard thanks to its powerful feature learning, but having a good representation goes further than achieving good performance on a target task \cite{bengio2013representation}. For example, TDA's stability can make deep representation resilient to input perturbation \cite{de2022ripsnet}. Below we review two categories of deep topological representations.
    
    
    \textit{Constrained Representations:} 
    One approach is to train deep neural networks to learn representations that preserve persistent homology of the input data.
    Again, TDA's versatility ensures the feasibility of this as the topological signature can be computed for both the input and the internal representation. For example, Topological Autoencoders~\cite{moor2020topological} does the alignment through a loss minimizing the divergence between input and latent representation topologies (both captured by PDs).

     \textit{Augmented Representations:} Another approach for topological representation is augmenting the deep features with topological signatures. Persistence Enhanced Graph Network (PEGN) \cite{Zhao2020} developed graph spatial convolution that builds on persistence homology. Normally, convolution filters are made adaptive to local graph structures by using node degree information. In contrast, PEGN weights the message passing (between nodes) by neighborhood information captured by persistence images. Moreover, Graph Filteration Learning (GFL) \cite{Hofer2020} adapts the readout operation (a graph pooling-like operation) in Graph Neural Network (GNN) to be topologically aware. BDs are computed for the graph nodes feature and vectorized. Interestingly, the filtration function is learned end-to-end. Topological Graph Layer (TOGL) \cite{Horn2022}  extends GFL's idea and learns multiple filtrations of a graph (rather than one) in an end-to-end manner.

    Unlike the embedding layers (e.g. PersLay~\cite{Carriere2019}) that expect a pre-specified input type (e.g. PDs), the topological representation layers discussed in this section enjoy more flexibility regarding the input and placement in the network. This comes with the attached cost of requiring careful design choices and guarantees on the layer characteristics (e.g. consistency of gradients in \cite{Hofer2020}).

\subsection{Topological Loss}
\label{sec:top_loss}
The most common approach for leveraging topology in deep learning is incorporating a topological penalty in the loss. The popularity of the approach stems from the fact that Loss-based integration is straightforward and doesn't require changing the architecture or adding additional layers. The only caveat is that the loss should differentiable and easy to compute.
As iterated previously the capability of topological features in capturing the complex structure of the data means deep learning can learn robust representations guided by the topological loss. Thus, the representations are likely invariant w.r.t typical transformations present in real-world datasets such as noise, and outliers. An example of this is a common persistence loss \cite{hu2019topology}, 
which minimizes the difference between a predicted persistence diagram $\mathrm{PD}_X$  and the true diagram $\mathrm{PD}_Y$: 

\begin{equation}
   \mathcal{L}_{\text{topological}} = d(\mathrm{PD}_X,\mathrm{PD}_Y) 
\end{equation}

This has been used either as a standalone loss or as a regularizer (i.e. augmenting another loss) \cite{hu2019topology} in applications such as semantic segmentation \cite{hu2019topology}, generative modeling \cite{wang2020topogan}.



As discussed in \ref{sec:tda_embedding}, PDs do not lend themselves to vector representations in Euclidean space. Moreover, the PD is not differentiable (a key requirement for using backpropagation).
One strategy to resolve this is leveraging a divergence or metric that can handle PDs. The $p$-Wasserstein\footnote{The ``Wasserstein'' distance in TDA literature is slightly different from the common Wasserstein (i.e.  Kantrovich optimal mass transport \cite{peyre2019computational}) metric. The first seeks a deterministic bijection that best aligns the diagrams (hard assignment) and the mass can be freely added to or removed from the diagonal. The latter is based on probabilistic coupling (soft assignment). This also has implications for the kind of algorithms that can be used to estimate the distance.
} distance and the bottleneck distance are popular choices: 
\begin{align}
    d_{p,q}(\mathrm{PD}_X,\mathrm{PD}_Y)&= \Big[ \inf_{\pi \in \Pi(\mathrm{PD}_X, \mathrm{PD}_Y) } \sum_{t \in \mathrm{PD}_X} \| t - \pi(t)\|_{q}^{p} \Big]^{\frac{1}{p}} \label{eq:p-wasserstein} \\ 
    d_{\infty}(\mathrm{PD}_X,\mathrm{PD}_Y)&= \inf_{\pi\in \Pi(\mathrm{PD}_X, \mathrm{PD}_Y) } \sup_{t \in \mathrm{PD}_X} \| t - \pi(t)\|_{\infty} \label{eq:bottleneck}
\end{align}
where $t$ is a point corresponding to a $(b_i, d_i)\in\mathbb{R}^2$ that is in $\mathrm{PD_X}$, and where
$\Pi(\mathrm{PD}_X, \mathrm{PD}_Y)$ denotes a the set of bijection between $\mathrm{PD}_X$ and $\mathrm{PD}_Y$, and $\|.\|_q$ is the $\ell_q$ Euclidean norm. 
It can be seen that bottleneck distance is the largest distance between any pair of corresponding points across all bijections that preserve the partial ordering of the points (i.e. we cannot match a point with a birth time greater than another point's death time). This ensures that the topological features to be matched are comparable. 

The initial popularity of bottleneck distance is perhaps fueled by a \textit{stability theorem} \cite{cohen2005stability} for PDs of continuous functions. 
According to this theorem, bottleneck distance is controlled by $L_\infty$ distance, that is 
\begin{equation}
d_{\infty}(\mathrm{PD}_{f_1},\mathrm{PD}_{f_2}) \leq C \| f_1-f_2\|_{\infty}
\end{equation}
form some constant $C$. In effect, this means that the diagrams are stable with respect to small perturbations of the underlying data. 
A similar stability result exists for the $p$-Wasserstein distance. 
These are the foundation of the stability guarantees by recent deep learning works such as the stability of Heat Kernel Signature in graphs \cite{Carriere2019} and stability of mini-batch-based diagram distances in Topological Autoencoders \cite{moor2020topological}. 
 

 Among the limitations of \eqref{eq:p-wasserstein} and \eqref{eq:bottleneck} is the high computational budget needed by these distances when the number of points is large. As the distance requires point-wise matching, the  computational complexity is $\mathcal{O}(n^3)$ for $n$ points \cite{anirudh2016riemannian}. Also, in many applications \cite{wang2020topogan,chen2019topological}, we aim to learn a model $f_w$ that aligns a predicted diagram $\mathrm{PD}_P$ with a target (i.e. ground truth)  diagram $\mathrm{PD}_T$ by gradually moving $\mathrm{PD}_P$ points towards $\mathrm{PD}_T$. This is typically achieved by pushing $w$ in the negative direction of $\nabla_w \mathcal{L}_{\text{topological}}$ and, obviously, assumes that the loss is differentiable w.r.t.~the diagram. While the Wasserstein distance satisfies this requirement in general, it can have some instability issues \cite{solomon2021fast}. Below, we select a few representative papers using topological losses in various applications and show how they handle these issues.

In generative modeling, TopoGAN \cite{wang2020topogan} uses a slightly modified 1-Wassertsein distance to align the diagrams of generated and real images in medical image applications. The loss ignores the death time and focuses only on the birth time of the diagram features. Framed in this way, the loss becomes similar to the Sliced Wasserstein \cite{peyre2019computational} which can be computed efficiently and is still differentiable. A similar loss was used by \cite{hu2019topology} for segmentation to encourage the deep model to produce output whose topology is close to the ground truth. The cross-entropy loss is augmented with the 2-Wasserstein loss between persistence diagrams. To alleviate the computational burden, the method performs the calculation on a single small image patch (part of the image) at a time. In \cite{Clough2022} the authors rely on  Betti numbers for semi-supervised image segmentation. A notable advantage here is the output of a network trained on a small set of labeled images can still capture the actual Betti numbers correctly. This gives us the opportunity to train the model initially on a small labeled dataset guided by the Betti numbers loss $\mathcal{L}_{\beta}$. Then, the model is fine-tuned using large unlabeled dataset and guided by a loss (that incorporates $\mathcal{L}_{\beta}$).  Since Betti numbers estimation is  robust for the unlabeled data, $\mathcal{L}_{\beta}$ will regularize the second stage of training (fine-tuning).
In classification, \cite{chen2019topological} uses a topological regularizer. To speed up the computation it focuses on the zero homological dimension where the persistence computations are particularly faster.

\subsection{Deep Topological Analytics}
\label{sec:tda-measure}



The complementary value of TDA goes beyond \textit{on-training integration} and constructing topological neural architectures. In fact, leveraging TDA methods \textit{post-training} can be even more insightful and powerful. Currently, researchers use TDA to address deep learning transparency ~\cite{Liu2020}, studying model complexity \cite{Rieck2019} and even tracking down answers for seemingly mysterious aspects of deep learning e.g. why deep networks outperform shallow ones \cite{naitzat2020topology}. These efforts are centered around analyzing deep models using TDA approaches. Hence, we call it \textit{deep topological analytics}. Due to space limitations, we explore only two aspects of it below.


     \textit{Quantifying structural complexity:}  \cite{Watanabe2021} treats the neural networks as a weighted graph $G(V, E)$ where $V$ and $E$ denote the network neurons and the relevance scores (computed from weights); respectively. By 
     computing persistence features (e.g. Betti numbers) across filtration, we can gain insight into the network complexity. For example, the increase in the Betti number (the occurrence of a cycle between a set of neurons) can reflect the complexity of knowledge in the deep neural networks. In \cite{Rieck2019} the authors follow the same line and further develop training optimization strategies (e.g. early stopping) informed by homological features.


    

    \textit{Visual exploration of models:} another use of TDA here is providing a post-hoc explanation and/or visual exploration of the internal functioning of deep models. For example, topological information provides insight into the overall structure of high-dimensional functions. The authors in \cite{Liu2020} use this to offer a scalable visual exploration tool for data-driven (and black box) models. This is an important research problem whose key challenge is doing it in an intuitive way. They also use {\em topological splines} 
     to visualize the high dimensional error landscape of the models. Similarly, TopoAct~\cite{Rathore2021} offers insightful information on neural network learned representations and provides a visual exploration tool to study topological summaries of activation vectors.

\section{Discussion} \label{sec:discussion}

 TDA is a steadily developing and promising area, with successes in a wide variety of applications. However, there are open questions in applying TDA with deep neural networks. In this section, we highlight several open challenges for future research of deep TDA in both practical and theoretical aspects and paint a speculative picture by outlining what persistent homology holds for the future. We also note some open-source implementations for researchers to get started.

\subsection{Challenges} \label{sec:challenges}
Despite the success of TDA and its use in deep learning we describe a few notable challenges here. 
   




\textbf{Computational cost:} Many aspects of calculating persistent homology are computational intractable. The construction of the \v{C}ech complex for a given $r$ is known to be an NP-hard task. Computing Betti numbers for a given simplicial complex are also infesable to compute for very large scale complexes.
Costs of calculating TDA information adds to the already computationally expensive deep learning routines.

\textbf{Lack of universal framework for vectorization:} There is no universally accepted framework for incorporating topological information into deep learning. 
 This is a theoretical matter as well as a computational one, for example there is a lack of strong theory encoding persistence diagrams as vectors, as discussed earlier. 
 There have been a variety of ad-hoc solutions of varying merit, recently cataloged in \cite{aliSurveyVectorizationMethods2022}. 
 Alternatively, vectorization methods have been chosen as part of learning strategies \cite{hofer2017deep,moor2020topological}.

\textbf{Statistical guarantees:} Through this article we have not discussed the statistical aspects of persistence due to finite sampling. For example, there is no guarantee that the PD derived from $X$ reflects the true homology of $\mathcal{M}$. 
The framework for understanding the statistical robustness of persistence information is evolving.
Some simple strategies for verification such as sub-sampling and cross-validation have been used in the literature \cite{chazal2021introduction}. 
There is scope to further understand issues such as the minimum number of data points required to guarantee robust PDs. 
Furthermore, persistence is not well understood from a probabilistic point of view, e.g.~the distribution of persistence from a distribution of shapes. 





\textbf{High-dimensional learning challenge:} There is no underlying theoretical framework for what topological features to expect with high-dimensional data. 
While abstract topological spaces can be enormously complex in high dimensions, we do not know whether to expect data to behave similarly.
Moreover high dimensional homological features are unatainable due to computational cost, and in any case sensitivity of PDs to sampling or noise is not well understood in high dimension.
This makes learning the underlying topology of the data for use in deep neural networks challenging. 




\textbf{The need for a good backpropagation strategy:} 
The differentiability of PDs or other homological quantities is not guaranteed or necessarily well understood.
This makes backpropagation in deep neural networks that incorporate
topological signatures extremely challenging or only feasible under special conditions~\cite{moor2020topological}.

\textbf{Capturing multi-variate persistence:}
In some cases, multiple concurrent filtrations are needed to fully capture the topology of the data manifold, especially for data in higher dimensions. 
This leads to \textit{multi-variate persistence}, where the birth and death of topological features occur in multiple dimensions.
This notion of persistence does not have a complete discrete invariant, unlike the one-dimensional BD that we've discussed so far. 
For the practical use of multi-variate persistence in deep learning, we would need new theoretical frameworks and better computational methods.



\subsection{Successes and Future Directions:}

Deep TDA has demonstrated potential in a variety of challenging settings. 
The invariance of PH information to continuous deformation means TDA applies well to settings where objects should have consistent shapes but may be transformed in some way.
TDA also performs well at bridging the gap between structural information and prior knowledge. 
If we have prior knowledge of the topology of a class of objects, then PDs are an effective tool for the classification and comparison of data against this class, even in the presence of noise or limited data.
This robustness incorporates well into deep learning.

TDA can produce good results in small datasets, this is especially useful for medical imaging applications where expense and privacy concerns limit data acquisition~\cite{Byrne2021}.
TDA has also been used in other settings with limited or noisy data such as 
power forecasting~\cite{Senekane2021}, segmenting aerial photography~\cite{Mosinska2018} and astronomy~\cite{Murugan2019}. 

In some applications, topological information may be more significant than statistical (pixel-wise) information.
For example, in~\cite{Vukicevic2017}
detecting holes between heart chambers is more important than inferring the thickness of septal walls. 
For these types of applications, a loss function combining topological and statistical information can be adjusted in favor of topology, when training a network.

As PH encapsulates global structure, developing topological loss functions could suppress small false positives or false negatives, particularly for computer vision tasks. 
For example, in image segmentation, morphological operations or conditional random field-based techniques are used to remove local errors, but they do not possess knowledge of global topology.
The benefit of PH-based loss is that the correct global topology can be propagated, with local label smoothness.




It would be interesting to explore sophisticated deep learning architectures that learn mappings between high dimensional data and their corresponding PDs or other topological representations, furthering \cite{de2022ripsnet}.
Moreover, deep learning may yet yield new kinds of topological representation other than PDs, with robustness to different data deformations.
PH could have further applications in multi-class open-set problems (where data may have unknown classes). If the topology among classes is relatively consistent, then the object labels of unknown classes can be better predicted.

\subsection{Implementations} \label{sec:tools}

There are a number of open-source implementations of TDA available to practitioners. Here we present two libraries that have interfaces with deep learning architectures.


GUDHI is an open-source library\footnote{\url{https://gudhi.inria.fr/}} that implements relevant geometric data structures and TDA algorithms, and it can be integrated into the TensorFlow framework. 
PersLay~\cite{Carriere2019} and RipsLayer are implementations using GUDHI that learn persistence representations from complexes and PDs. They can handle automatic differentiation and are readily integrated in deep learning architectures.


Giotto-deep\footnote{\url{https://github.com/giotto-ai/giotto-deep}} is an open-source extension of the Giotto-TDA library. It aims to provide seamless integration between TDA and deep learning on top of PyTorch. 
To use topology for both pre-processing data (using a variety of available methods) and using it within neural networks, the developers aim to provide several off-the-shelf architectures. 
One such example is that of Persformer~\cite{Reinauer2021}.


\section{Conclusion} \label{sec:conclusion}

    The recent growth in TDA, and the established efficacy of deep learning, has meant that integration of these techniques has been inevitable.
    There is no universal paradigm for combining TDA and deep learning.
    This article surveyed numerous ways in which these frameworks have benefited each other. We began with an overview of the key TDA concepts. 
    Following this we reviewed TDA in deep learning from a variety of perspectives.
    We described numerous challenges and opportunities that remain in this field, as well as some observed success.


\bibliographystyle{named}
\bibliography{refs}

\begin{thebibliography}{}

\bibitem[\protect\citeauthoryear{Adams \bgroup \em et al.\egroup
  }{2017}]{Adams2017}
Henry Adams, Tegan Emerson, Michael Kirby, Rachel Neville, Chris Peterson,
  Patrick Shipman, Sofya Chepushtanova, Eric Hanson, Francis Motta, and Lori
  Ziegelmeier.
\newblock Persistence images: A stable vector representation of persistent
  homology.
\newblock {\em J. Mach. Learn. Res.}, 18(1):218–252, jan 2017.

\bibitem[\protect\citeauthoryear{Ali \bgroup \em et al.\egroup
  }{2022}]{aliSurveyVectorizationMethods2022}
Dashti Ali, Aras Asaad, Maria-Jose Jimenez, Vidit Nanda, Eduardo
  {Paluzo-Hidalgo}, and Manuel {Soriano-Trigueros}.
\newblock A survey of vectorization methods in topological data analysis,
  December 2022.

\bibitem[\protect\citeauthoryear{Anirudh \bgroup \em et al.\egroup
  }{2016}]{anirudh2016riemannian}
Rushil Anirudh, Vinay Venkataraman, Karthikeyan Natesan~Ramamurthy, and Pavan
  Turaga.
\newblock A riemannian framework for statistical analysis of topological
  persistence diagrams.
\newblock In {\em Proceedings of the IEEE conference on computer vision and
  pattern recognition workshops}, pages 68--76, 2016.

\bibitem[\protect\citeauthoryear{Bengio \bgroup \em et al.\egroup
  }{2013}]{bengio2013representation}
Yoshua Bengio, Aaron Courville, and Pascal Vincent.
\newblock Representation learning: A review and new perspectives.
\newblock {\em IEEE transactions on pattern analysis and machine intelligence},
  35(8):1798--1828, 2013.

\bibitem[\protect\citeauthoryear{Bubenik \bgroup \em et al.\egroup
  }{2017}]{Bubenik2017}
Peter Bubenik, Vin de~Silva, and Jonathan~A. Scott.
\newblock Interleaving and gromov-hausdorff distance.
\newblock {\em arXiv: Category Theory}, 2017.

\bibitem[\protect\citeauthoryear{Bubenik}{2020}]{Bubenik2020}
Peter Bubenik.
\newblock The persistence landscape and some of its properties.
\newblock In {\em Topological Data Analysis}, pages 97--117. Springer
  International Publishing, 2020.

\bibitem[\protect\citeauthoryear{Byrne \bgroup \em et al.\egroup
  }{2021}]{Byrne2021}
Nick Byrne, James~R. Clough, Giovanni Montana, and Andrew~P. King.
\newblock A persistent homology-based topological loss function for multi-class
  {CNN} segmentation of cardiac {MRI}.
\newblock In {\em Statistical Atlases and Computational Models of the Heart.},
  pages 3--13. Springer International Publishing, 2021.

\bibitem[\protect\citeauthoryear{Cang and Wei}{2017}]{Cang2017}
Zixuan Cang and Guo-Wei Wei.
\newblock {TopologyNet}: Topology based deep convolutional and multi-task
  neural networks for biomolecular property predictions.
\newblock {\em {PLOS} Computational Biology}, 13(7), jul 2017.

\bibitem[\protect\citeauthoryear{Cang \bgroup \em et al.\egroup
  }{2015}]{Cang2015}
Zixuan Cang, Lin Mu, Kedi Wu, Kristopher Opron, Kelin Xia, and Guo-Wei Wei.
\newblock A topological approach for protein classification.
\newblock {\em Molecular Based Mathematical Biology}, 3(1):null, 2015.

\bibitem[\protect\citeauthoryear{Carlsson}{2009}]{carlssonTopologyData2009}
Gunnar Carlsson.
\newblock Topology and data.
\newblock {\em Bulletin of the American Mathematical Society}, 46(2):255--308,
  January 2009.

\bibitem[\protect\citeauthoryear{Carri{\`e}re \bgroup \em et al.\egroup
  }{2019}]{Carriere2019}
Mathieu Carri{\`e}re, Fr{\'e}d{\'e}ric Chazal, Yuichi Ike, Th{\'e}o Lacombe,
  Martin Royer, and Yuhei Umeda.
\newblock Perslay: A neural network layer for persistence diagrams and new
  graph topological signatures.
\newblock In {\em International Conference on Artificial Intelligence and
  Statistics}, 2019.

\bibitem[\protect\citeauthoryear{Chazal and
  Michel}{2021}]{chazal2021introduction}
Fr{\'e}d{\'e}ric Chazal and Bertrand Michel.
\newblock An introduction to topological data analysis: fundamental and
  practical aspects for data scientists.
\newblock {\em Frontiers in artificial intelligence}, 4, 2021.

\bibitem[\protect\citeauthoryear{Chen \bgroup \em et al.\egroup
  }{2019}]{chen2019topological}
Chao Chen, Xiuyan Ni, Qinxun Bai, and Yusu Wang.
\newblock A topological regularizer for classifiers via persistent homology.
\newblock In {\em The 22nd International Conference on Artificial Intelligence
  and Statistics}, pages 2573--2582. PMLR, 2019.

\bibitem[\protect\citeauthoryear{Chevyrev \bgroup \em et al.\egroup
  }{2020}]{Chevyrev2020}
Ilya Chevyrev, Vidit Nanda, and Harald Oberhauser.
\newblock Persistence paths and signature features in topological data
  analysis.
\newblock {\em {IEEE} Transactions on Pattern Analysis and Machine
  Intelligence}, 42(1):192--202, jan 2020.

\bibitem[\protect\citeauthoryear{Clough \bgroup \em et al.\egroup
  }{2022}]{Clough2022}
James~R. Clough, Nicholas Byrne, Ilkay Oksuz, Veronika~A. Zimmer, Julia~A.
  Schnabel, and Andrew~P. King.
\newblock A topological loss function for deep-learning based image
  segmentation using persistent homology.
\newblock {\em {IEEE} Transactions on Pattern Analysis and Machine
  Intelligence}, 44(12):8766--8778, dec 2022.

\bibitem[\protect\citeauthoryear{Cohen-Steiner \bgroup \em et al.\egroup
  }{2005}]{cohen2005stability}
David Cohen-Steiner, Herbert Edelsbrunner, and John Harer.
\newblock Stability of persistence diagrams.
\newblock In {\em Proceedings of the twenty-first annual symposium on
  Computational geometry}, pages 263--271, 2005.

\bibitem[\protect\citeauthoryear{de Surrel \bgroup \em et al.\egroup
  }{2022}]{de2022ripsnet}
Thibault de~Surrel, Felix Hensel, Mathieu Carri{\`e}re, Th{\'e}o Lacombe,
  Yuichi Ike, Hiroaki Kurihara, Marc Glisse, and Fr{\'e}d{\'e}ric Chazal.
\newblock Ripsnet: a general architecture for fast and robust estimation of the
  persistent homology of point clouds.
\newblock In {\em Topological, Algebraic and Geometric Learning Workshops
  2022}, pages 96--106. PMLR, 2022.

\bibitem[\protect\citeauthoryear{{Edelsbrunner} \bgroup \em et al.\egroup
  }{2002}]{edelsbrunnerTopologicalPersistenceSimplification2002}
{Edelsbrunner}, {Letscher}, and {Zomorodian}.
\newblock Topological persistence and simplification.
\newblock {\em Discrete \& Computational Geometry}, 28(4):511--533, November
  2002.

\bibitem[\protect\citeauthoryear{Hofer \bgroup \em et al.\egroup
  }{2017}]{hofer2017deep}
Christoph Hofer, Roland Kwitt, Marc Niethammer, and Andreas Uhl.
\newblock Deep learning with topological signatures.
\newblock {\em Advances in neural information processing systems}, 30, 2017.

\bibitem[\protect\citeauthoryear{Hofer \bgroup \em et al.\egroup
  }{2020}]{Hofer2020}
Christoph Hofer, Florian Graf, Bastian Rieck, Marc Niethammer, and Roland
  Kwitt.
\newblock Graph filtration learning.
\newblock In Hal~Daumé III and Aarti Singh, editors, {\em Proceedings of the
  37th International Conference on Machine Learning}, volume 119 of {\em
  Proceedings of Machine Learning Research}, pages 4314--4323. PMLR, 13--18 Jul
  2020.

\bibitem[\protect\citeauthoryear{Horn \bgroup \em et al.\egroup
  }{2022}]{Horn2022}
Max Horn, Edward~De Brouwer, Michael Moor, Yves Moreau, Bastian Rieck, and
  Karsten Borgwardt.
\newblock Topological graph neural networks.
\newblock In {\em International Conference on Learning Representations}, 2022.

\bibitem[\protect\citeauthoryear{Hu \bgroup \em et al.\egroup
  }{2019}]{hu2019topology}
Xiaoling Hu, Fuxin Li, Dimitris Samaras, and Chao Chen.
\newblock Topology-preserving deep image segmentation.
\newblock {\em Advances in neural information processing systems}, 32, 2019.

\bibitem[\protect\citeauthoryear{Kali{\v{s}}nik}{2018}]{Kalisnik2018}
Sara Kali{\v{s}}nik.
\newblock Tropical coordinates on the space of persistence barcodes.
\newblock {\em Foundations of Computational Mathematics}, 19(1):101--129, jan
  2018.

\bibitem[\protect\citeauthoryear{Kim \bgroup \em et al.\egroup
  }{2020}]{Kim2020}
K.~Kim, J.~Kim, M.~Zaheer, J.~Kim, F.~Chazal, and L.~andWasserman.
\newblock Pllay: Efficient topological layer based on persistent landscapes.
\newblock In {\em In Advances in Neural Information Processing Systems (NeurIPS
  2020)}, 2020.

\bibitem[\protect\citeauthoryear{Kwitt \bgroup \em et al.\egroup
  }{2015}]{Kwitt2015}
Roland Kwitt, Stefan Huber, Marc Niethammer, Weili Lin, and Ulrich Bauer.
\newblock Statistical topological data analysis - a kernel perspective.
\newblock In C.~Cortes, N.~Lawrence, D.~Lee, M.~Sugiyama, and R.~Garnett,
  editors, {\em Advances in Neural Information Processing Systems}, volume~28.
  Curran Associates, Inc., 2015.

\bibitem[\protect\citeauthoryear{Liu \bgroup \em et al.\egroup
  }{2020}]{Liu2020}
Shusen Liu, Jim Gaffney, Luc Peterson, Peter~B. Robinson, Harsh Bhatia, Valerio
  Pascucci, Brian~K. Spears, Peer-Timo Bremer, Di~Wang, Dan Maljovec, Rushil
  Anirudh, Jayaraman~J. Thiagarajan, Sam~Ade Jacobs, Brian C.~Van Essen, David
  Hysom, and Jae-Seung Yeom.
\newblock Scalable topological data analysis and visualization for evaluating
  data-driven models in scientific applications.
\newblock {\em {IEEE} Transactions on Visualization and Computer Graphics},
  26(1):291--300, jan 2020.

\bibitem[\protect\citeauthoryear{Mileyko \bgroup \em et al.\egroup
  }{2011}]{Mileyko2011}
Yuriy Mileyko, Sayan Mukherjee, and John Harer.
\newblock Probability measures on the space of persistence diagrams.
\newblock {\em Inverse Problems}, 27(12):124007, nov 2011.

\bibitem[\protect\citeauthoryear{Moor \bgroup \em et al.\egroup
  }{2020}]{moor2020topological}
Michael Moor, Max Horn, Bastian Rieck, and Karsten Borgwardt.
\newblock Topological autoencoders.
\newblock In {\em International conference on machine learning}, pages
  7045--7054. PMLR, 2020.

\bibitem[\protect\citeauthoryear{Mosinska \bgroup \em et al.\egroup
  }{2018}]{Mosinska2018}
Agata Mosinska, Pablo Marquez-Neila, Mateusz Kozinski, and Pascal Fua.
\newblock Beyond the pixel-wise loss for topology-aware delineation.
\newblock In {\em 2018 {IEEE}/{CVF} Conference on Computer Vision and Pattern
  Recognition}. {IEEE}, jun 2018.

\bibitem[\protect\citeauthoryear{Murugan and Robertson}{2019}]{Murugan2019}
Jeff Murugan and Duncan Robertson.
\newblock An introduction to topological data analysis for physicists: From lgm
  to frbs, 2019.

\bibitem[\protect\citeauthoryear{Naitzat \bgroup \em et al.\egroup
  }{2020}]{naitzat2020topology}
Gregory Naitzat, Andrey Zhitnikov, and Lek-Heng Lim.
\newblock Topology of deep neural networks.
\newblock {\em The Journal of Machine Learning Research}, 21(1):7503--7542,
  2020.

\bibitem[\protect\citeauthoryear{Peyr{\'e} \bgroup \em et al.\egroup
  }{2019}]{peyre2019computational}
Gabriel Peyr{\'e}, Marco Cuturi, et~al.
\newblock Computational optimal transport: With applications to data science.
\newblock {\em Foundations and Trends{\textregistered} in Machine Learning},
  11(5-6):355--607, 2019.

\bibitem[\protect\citeauthoryear{Pun \bgroup \em et al.\egroup
  }{2022}]{Pun2022}
Chi~Seng Pun, Si~Xian Lee, and Kelin Xia.
\newblock Persistent-homology-based machine learning: a survey and a
  comparative study.
\newblock {\em Artificial Intelligence Review}, 55(7):5169--5213, feb 2022.

\bibitem[\protect\citeauthoryear{Rathore \bgroup \em et al.\egroup
  }{2021}]{Rathore2021}
Archit Rathore, Nithin Chalapathi, Sourabh Palande, and Bei Wang.
\newblock {TopoAct}: Visually exploring the shape of activations in deep
  learning.
\newblock {\em Computer Graphics Forum}, 40(1):382--397, jan 2021.

\bibitem[\protect\citeauthoryear{Reinauer \bgroup \em et al.\egroup
  }{2021}]{Reinauer2021}
Raphael Reinauer, Matteo Caorsi, and Nicolas Berkouk.
\newblock Persformer: A transformer architecture for topological machine
  learning, 2021.

\bibitem[\protect\citeauthoryear{Rieck \bgroup \em et al.\egroup
  }{2019}]{Rieck2019}
Bastian Rieck, Matteo Togninalli, Christian Bock, Michael Moor, Max Horn,
  Thomas Gumbsch, and Karsten Borgwardt.
\newblock Neural persistence: {A} complexity measure for deep neural networks
  using algebraic topology.
\newblock In {\em International Conference on Learning Representations~(ICLR)},
  2019.

\bibitem[\protect\citeauthoryear{Senekane \bgroup \em et al.\egroup
  }{2021}]{Senekane2021}
Makhamisa Senekane, Naleli~Jubert Matjelo, and Benedict~Molibeli Taele.
\newblock Improving short-term output power forecasting using topological data
  analysis and machine learning.
\newblock In {\em 2021 International Conference on Electrical, Computer and
  Energy Technologies ({ICECET})}. {IEEE}, dec 2021.

\bibitem[\protect\citeauthoryear{Solomon \bgroup \em et al.\egroup
  }{2021}]{solomon2021fast}
Yitzchak Solomon, Alexander Wagner, and Paul Bendich.
\newblock A fast and robust method for global topological functional
  optimization.
\newblock In {\em International Conference on Artificial Intelligence and
  Statistics}, pages 109--117. PMLR, 2021.

\bibitem[\protect\citeauthoryear{Vukicevic \bgroup \em et al.\egroup
  }{2017}]{Vukicevic2017}
Marija Vukicevic, Bobak Mosadegh, James~K. Min, and Stephen~H. Little.
\newblock Cardiac 3d printing and its future directions.
\newblock {\em {JACC}: Cardiovascular Imaging}, 10(2):171--184, feb 2017.

\bibitem[\protect\citeauthoryear{Wang \bgroup \em et al.\egroup
  }{2020}]{wang2020topogan}
Fan Wang, Huidong Liu, Dimitris Samaras, and Chao Chen.
\newblock Topogan: A topology-aware generative adversarial network.
\newblock In {\em European Conference on Computer Vision}, pages 118--136.
  Springer, 2020.

\bibitem[\protect\citeauthoryear{Watanabe and Yamana}{2021}]{Watanabe2021}
Satoru Watanabe and Hayato Yamana.
\newblock Topological measurement of deep neural networks using persistent
  homology.
\newblock {\em Annals of Mathematics and Artificial Intelligence},
  90(1):75--92, jul 2021.

\bibitem[\protect\citeauthoryear{Xia \bgroup \em et al.\egroup
  }{2017}]{Xia2017}
Kelin Xia, Zhiming Li, and Lin Mu.
\newblock Multiscale persistent functions for biomolecular structure
  characterization.
\newblock {\em Bulletin of Mathematical Biology}, 80(1):1--31, nov 2017.

\bibitem[\protect\citeauthoryear{Zaheer \bgroup \em et al.\egroup
  }{2017}]{Zaheer2017}
Manzil Zaheer, Satwik Kottur, Siamak Ravanbakhsh, Barnabas Poczos, Ruslan~R
  Salakhutdinov, and Alexander~J Smola.
\newblock Deep sets.
\newblock In {\em Advances in Neural Information Processing Systems 30}, pages
  3391--3401, 2017.

\bibitem[\protect\citeauthoryear{Zhao \bgroup \em et al.\egroup
  }{2020}]{Zhao2020}
Qi~Zhao, Ze~Ye, Chao Chen, and Yusu Wang.
\newblock Persistence enhanced graph neural network.
\newblock In Silvia Chiappa and Roberto Calandra, editors, {\em Proceedings of
  the Twenty Third International Conference on Artificial Intelligence and
  Statistics}, volume 108 of {\em Proceedings of Machine Learning Research},
  pages 2896--2906. PMLR, 26--28 Aug 2020.

\bibitem[\protect\citeauthoryear{Zhu \bgroup \em et al.\egroup
  }{2016}]{Zhu2016}
Xiaojin Zhu, A.~Vartanian, Mani Bansal, Duy Nguyen, and Luke Brandl.
\newblock Stochastic multiresolution persistent homology kernel.
\newblock In {\em International Joint Conference on Artificial Intelligence},
  2016.

\bibitem[\protect\citeauthoryear{Zieli{\'{n}}ski \bgroup \em et al.\egroup
  }{2020}]{Zielinski2020}
Bartosz Zieli{\'{n}}ski, Micha{\l} Lipi{\'{n}}ski, Mateusz Juda, Matthias
  Zeppelzauer, and Pawe{\l} D{\l}otko.
\newblock Persistence codebooks for topological data analysis.
\newblock {\em Artificial Intelligence Review}, 54(3):1969--2009, sep 2020.

\bibitem[\protect\citeauthoryear{Zomorodian}{2010}]{Zomorodian2010}
Afra Zomorodian.
\newblock Fast construction of the vietoris-rips complex.
\newblock {\em Computers and Graphics}, 34(3):263--271, jun 2010.

\end{thebibliography}
\end{document}